\documentclass[10pt,twocolumn,letterpaper]{article}

\usepackage{iccv}
\usepackage{times}
\usepackage{epsfig}
\usepackage{graphicx}
\usepackage{amsmath}
\usepackage{amssymb}
\usepackage{array}
\usepackage{multirow}
\newcolumntype{M}[1]{>{\centering\arraybackslash}m{#1}}
\newcolumntype{N}{@{}m{0pt}@{}}

\graphicspath{ {./figures/} }

\usepackage[pagebackref=true,breaklinks=true,letterpaper=true,colorlinks,bookmarks=false]{hyperref}

\iccvfinalcopy 

\def\iccvPaperID{13} 

\ificcvfinal\fi

\begin{document}

\title{Enlisting 3D Crop Models and GANs for More Data Efficient and Generalizable Fruit Detection}

\author{Zhenghao Fei \quad Alex Olenskyj \quad Brian N. Bailey \quad Mason Earles\\
University of California, Davis\\
{\tt\small \{zfei, agolenskyj, bnbailey, jmearles\}@ucdavis.edu}
}

\maketitle
\ificcvfinal\thispagestyle{empty}\fi

\begin{abstract}
    Training real-world neural network models to achieve high performance and generalizability typically requires a substantial amount of labeled data, spanning a broad range of variation. This data-labeling process can be both labor and cost intensive. To achieve desirable predictive performance, a trained model is typically applied into a domain where the data distribution is similar to the training dataset. However, for many agricultural machine learning problems, training datasets are collected at a specific location, during a specific period in time of the growing season. Since agricultural systems exhibit substantial variability in terms of crop type, cultivar, management, seasonal growth dynamics, lighting condition, sensor type, etc, a model trained from one dataset often does not generalize well across domains. To enable more data efficient and generalizable neural network models in agriculture, we propose a method that generates photorealistic agricultural images from a synthetic 3D crop model domain into real world crop domains. The method uses a semantically constrained GAN (generative adversarial network) to preserve the fruit position and geometry. We observe that a baseline CycleGAN method generates visually realistic target domain images but does not preserve fruit position information while our method maintains fruit positions well. Image generation results in vineyard grape day and night images show the visual outputs of our network are much better compared to a baseline network. Incremental training experiments in vineyard grape detection tasks show that the images generated from our method can significantly speed the domain adaption process, increase performance for a given number of labeled images (i.e. data efficiency), and decrease labeling requirements.
\end{abstract}

\begin{figure}[t]
\begin{center}
\includegraphics[width=1.0\linewidth]{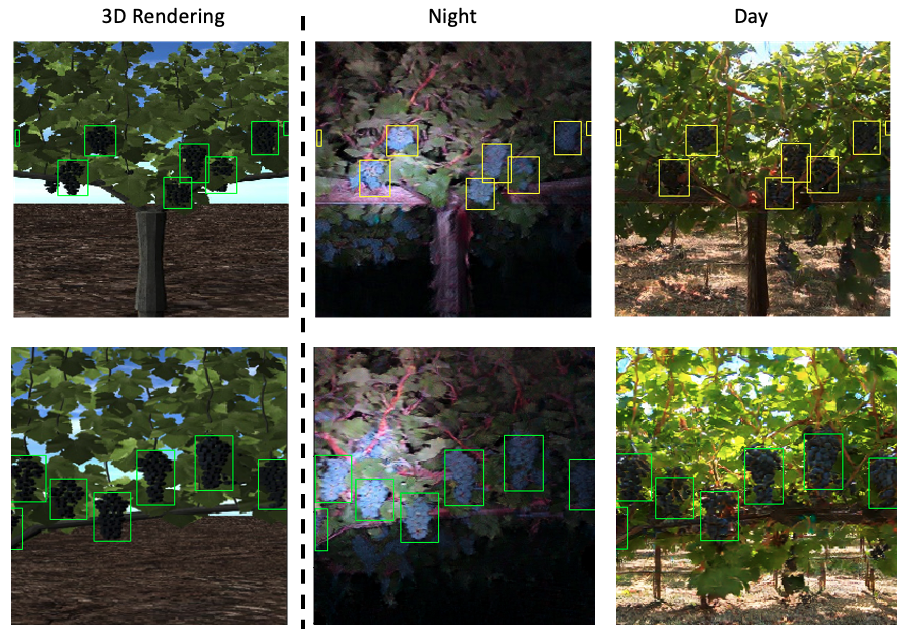}
   \caption{Given a synthetic 3D rendered grape image, our task aware semantically constrained GAN can generate various target real-world domain images with consistent label item locations.}
\label{fig:img-transfer}
\end{center}
\end{figure}

\section{Introduction}

Showing promising detection results in complex environments, deep neural network-based models (primarily convolutional neural networks, or CNNs) have been widely applied in agricultural applications. Bargoti and Underwood \cite{Bargoti2017} introduced Faster R-CNN \cite{Ren2015} for agricultural applications such as fruit detection in orchards, including mangoes, apples, and almonds. They showed that deep neural network based approaches achieved high accuracy in fruit detection. Santos \etal \cite{Santos2020} applied deep neural networks, YOLO (Redmon \etal \cite{Redmon2016}) for grape detection and Mask R-CNN (He \etal \cite{He2020}) for grape instance segmentation, to recognize and track grapes in RGB imagery. As another example, Zabawa \etal \cite{Zabawa2020} proposed an encoder-decoder semantic segmentation network for accurate and efficient grape counting. Vasconez \etal \cite{Vasconez2020} did a comprehensive evaluation of different CNNs applied to fruit detection and counting. Deep neural network based approaches have not only been applied to fruit detection and counting, but they have also been widely used in field-based robotic and automation applications, such as thinning, pruning, and harvesting. For instance, Zhang \etal \cite{Zhang2021} tested the use of CNNs in identification of tree trunks and branches for automated shake‐and‐catch apple harvesting. Majeed \etal \cite{Majeed2021} developed and tested CNNs for vine cordon detection to provide a reference for robotic green shoot thinning. 

While deep neural networks have been widely applied to various agricultural tasks, such approaches typically need a large amount of data to train high performing models. Yet, in the agricultural domain, publicly available datasets are very limited, and at the same time the data are usually very specific to the application scenario, plant variety, horticultural practice, lighting condition, seasons, and even camera type. Silwal \etal \cite{Silwal2021} showed that apple images captured at the same location can vary substantially given different camera systems, which affects model performance. If a model trained on data from one field does not work well when applied to data from a different field, or a model trained during one year does not work well for the next year, the resulting model will not be scalable. Thus, it is important to develop new techniques that enable successful adaptation of a deep learning model trained in one agricultural domain into a new domain (same crop but different horticultural practice, lighting condition, season, or camera type), while minimizing the amount of additional labeling required. 

To enable more data efficient and generalizable neural network models in agricultural applications, we propose a method that generates photorealistic agricultural images from a synthetic 3D crop model domain into real-world crop domains. The main contributions of our work include:

1.	A task-aware, semantically constrained GAN that translates images from one agricultural domain into another domain while keeping the task-related semantics (such as fruit position and size as in Figure \ref{fig:img-transfer}).

2.	A domain adaptation pipeline that improves model performance in another domain, both utilizing fine-tuning and semantically constrained GAN generated labeled images with a small number of labeled images in the target domain.

3.	Utilization of a 3D crop model to generate synthetic grape images for pre-training the grape detection model, and also using these synthetic images to generate photorealistic images with the same labels. This ultimately enables generation of unlimited free “labeled” images in the target domain.

\section{Related Work}
GANs \cite{Goodfellow2014} are generative models that are widely used for generating artificial new data (image) with the same distribution as the training data. The artificial images generated using GAN are visually realistic \cite{Karras2018}. Domain adaptation using GAN has gained a lot of attention in recent years. Zhu \etal \cite{Zhu2017} proposed an unpaired image-to-image translation method using cycle consistent adversarial networks (i.e., CycleGAN) to translate images from one domain to another without the need for paired image training data. Their method showed promising results in collection style transfer, object transfiguration, season transfer, and photo enhancement. However, CycleGAN does not specifically constrain the semantics of an image after translation. Consequently, the translated image often closely matches the general visual distribution of the original image, but objects within the image are often not well aligned. The idea of using GAN for domain adaptation has also been introduced into the agricultural domain. Giuffrida \etal \cite{Giuffrida2019} used adversarial unsupervised domain adaptation to reduce the domain shift between two datasets. They used an adversarial loss to match the statistics of the image features between two datasets without generating visually translated images. Moreover, the leaf counting dataset they used was derived from images taken in a controlled environment, as opposed to the field. Marino \etal \cite{Marino2020} applied CoGAN (Liu and Tuzel, \cite{Liu2016}) to bridge the domain gap between potato defect classification datasets. They tested their method on artificially brightened and different colored potatoes, achieving performance improvements compared to when no domain adaptation was applied. However, the domain gap and scene complexity in their experiments were relatively limited. Bellocchio \etal \cite{Bellocchio2020} combined an unsupervised domain adaptation network (i.e., CycleGAN) and a weakly-supervised fruit counting model to count fruits in four different orchards. The results show that their proposed approach is more accurate than the supervised baseline method alone, but due to the weakly-supervised fruit counting model, their method is limited to counting tasks. Gogoll \etal \cite{Gogoll2020} designed an unsupervised semantically consistent domain transfer method for plant/weed pixel-wise classification in new field environments. They utilized the idea that the image before and after transfer should have the same labels, which was enforced in the loss function when co-training the generators and target domain fully convolutional networks (FCNs) semantic segmentation model. They achieved very promising transfer results in the plant/weed classification task, and their method does not rely on any target domain labeled data. However, their method cannot explicitly avoid the ``trap" in which the task network and the generators cooperate to find a shortcut to trick the loss (e.g., the generator transfers plants to stones and the task network classifies stones as plants). Drees \etal \cite{Drees2021} extends the idea of using GAN to generate temporal predictions of plant growth in which the model learns from a plant growth model and produces realistic, reliable images of future growth stages of plants. Kierdorf \etal \cite{Kierdorf2021} proposed the use of conditional GAN for estimation of grapevine berries occluded by leaves by treating the occluded and non-occluded grapevine images as two domains, based on different leaf distributions, that can be translated to each other.

\section{Approach}
\subsection{Problem definition}

\begin{figure}[h]
\begin{center}
\includegraphics[width=1.0\linewidth]{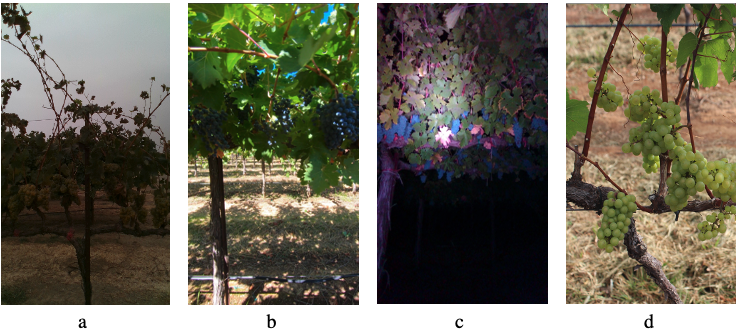}
\end{center}
   \caption{Images from four domains within grape vineyards. a) image collected at vineyard A on a shady day using an Intel RealSense D435i camera; b) images collected at vineyard B on a sunny day using a GoPro Hero7 Black camera; c) image collected at vineyard B during the night using an Intel RealSense D435i camera; d) image from the WGISD dataset \cite{Santos2020}.} 
\label{fig:multi-domain}
\end{figure}

For each domain adaptation task in our problem, there are two domains. The first is source domain A.  A is usually a well-labeled real-world dataset or a synthetically generated dataset with ground truth labels produced via a 3D rendering engine (i.e., a 3D crop model). The second domain is called the target domain $B$ . B refers to the domain where the model is applied for prediction. There are many images in domain $B$ , but a lack of ground-truth labels. The target domain $B$ and the source domain $A$ can be different in style but should have similar contexts. Specifically in agricultural applications, two domains should contain the same crop while having a difference in crop variety, lighting condition, the camera view distance/angle, management practice, etc. Figure \ref{fig:multi-domain} provides an example of different domain data for grape production in which the differences between domains is apparent.

In this study, our objective is to utilize the well-labeled source domain $A$ while using as little labeled data as possible in the target domain $B$ to train a task model \(T\) that can perform well in $B$. To achieve this, we want to learn a mapping between $A$ and $B$ given $N$ training samples \({{\{x_i^A\}}}_{i=1}^N \) where \(x_i^A\in A\) with task labels \(y_i^A\in Y^A\) and $M$ training samples \({{\{x_j^B\}}}_{j=1}^M\). Among $M$ training images in domain $B$, a few of them (\(k\)) can be labeled. The mapping from $A$ to $B$ is called \(G_{A}\) and the mapping from $B$ to $A$ is called \(G_{B}\). The generated image \(G_A(x_i^A)\)  should have the same visual style as images in domain $B$ but also maintain the same task-related semantics as the original image \(x_i\). We hypothesize that fine-tuning a task model \(T^B\) on the generated image \(G_A(x_i^A)\) should improve the performance of \(T^B\) in domain $B$.

\subsection{Task}

The main objective of our method is not only photorealistic image generation but also to utilize the generated images to facilitate domain adaptation. As a result, the machine learning task and corresponding task model $T$ is very important to our method. Generally speaking, the task model should be a fully differentiable model that can provide guidance through backpropagation to the image generation network. Here in this study, we chose object detection as our task as it is one of the more popular and important tasks in agricultural machine learning applications. Specifically, the object detection model we used in this work is YOLOv3 (Redmon and Farhadi, \cite{Redmon2018}).

\subsection{Method}

\begin{figure}[h]
\begin{center}
   \includegraphics[width=1.0\linewidth]{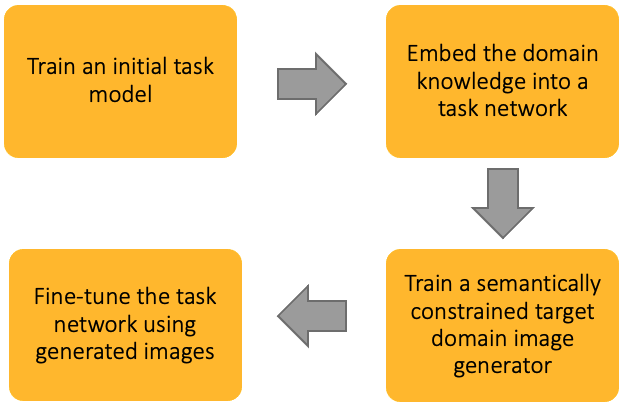}
\end{center}
   \caption{Training pipeline overview. The method includes four steps: 1) Train a initial task model using synthetic labeled data. 2) Fine-tune the pre-trained task model using a few labeled target domain images to embed the domain knowledge into the task model. 3) Train a semantically constrained GAN. 4) Fine-tune the task model using GAN generated images and labels. }
\label{fig:pipeline}
\end{figure}

In this problem, we assume that we have access to labeled source images \({{\{x_i^A\}}}_{i=1}^N\) with all task labels \(y_i^{A}\in{Y}^{A}\), \(M-k\)\((M > k)\) unlabeled training images  \({{\{x_j^{B}\}}}_{j=1}^{M-k}\), and $k$ labeled training images \({{\{x_j^{B}\}}}_{j=M-k+1}^M\) with  \(y_i^{B}\in{Y}^{B}\). We want to train an accurate task model \(T^B\) on domain $B$ using as few labeled training images as possible. The step-by-step method of doing this is shown below and an overview is shown in Figure \ref{fig:pipeline}.

\subsubsection{Train an initial task model}
Given the labeled source images \({{\{x_i^{A}\}}}_{i=1}^N\) with all task labels \(y_i^{A}\in{Y}^{A}\), we can train an initial task model \(T^A\) using the given data in a supervised way by minimizing the task loss \(L_{task}\). This model \(T^A\) performs well in domain $A$ but it performs relatively poorly in domain $B$ , and the level of performance degeneration is related to the domain gap between A and B (e.g., a model trained in a daylight domain usually performs better in another daylight domain than in a night domain).

\subsubsection{Embed the domain knowledge into a task network}
One of the main ideas behind our method is to embed the domain knowledge into a task network by fine-tuning the initial task model \(T^A\) using a few ($k$) labeled images in domain $B$ , the fine-tuned task model is named \(T^B\). Based on our finding, even fine-tuning with a single labeled image in domain $B$  can make \(T^B\) perform much better than \(T^A\). We call this step ``domain knowledge embedding".

\begin{figure}[h]
\begin{center}
   \includegraphics[width=1.0\linewidth]{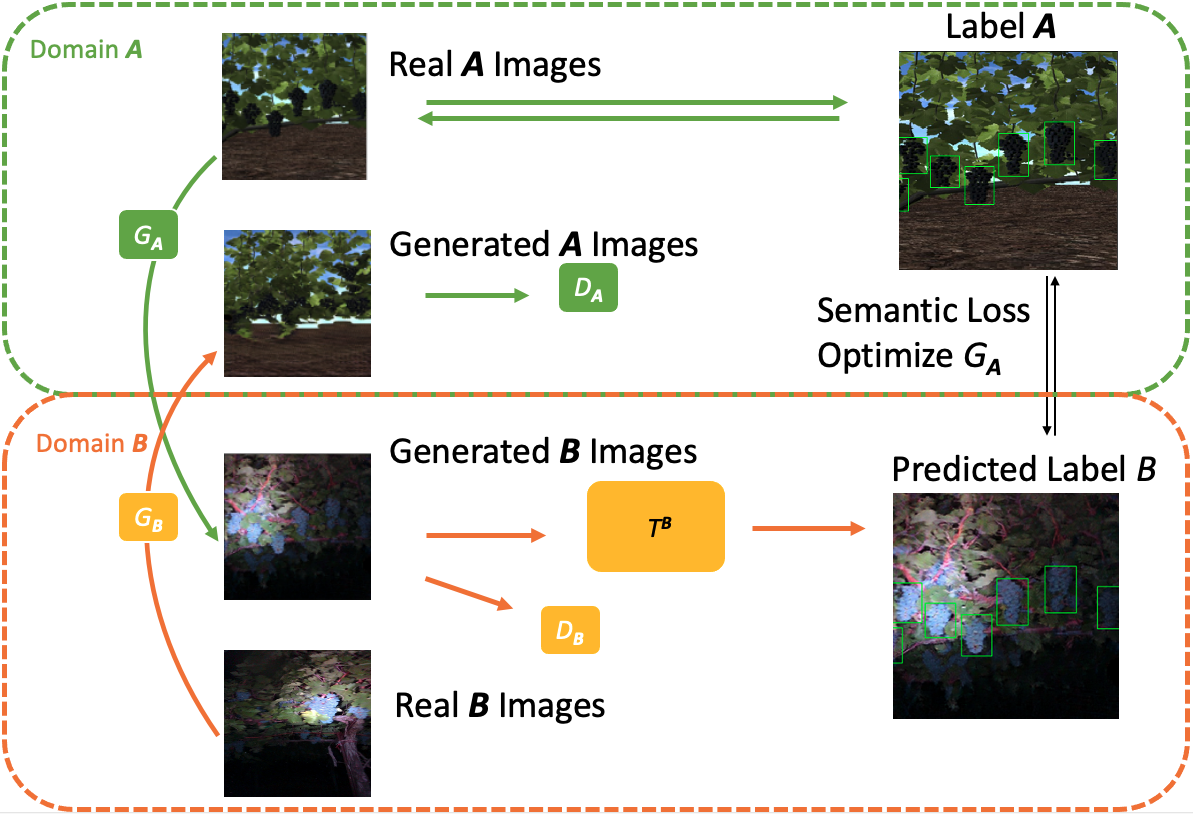}
\end{center}
   \caption{Overview of the proposed image generation network.}
\label{fig:network}
\end{figure}

\subsubsection{Train a semantically constrained target domain image generator}
To achieve the generation of images from domain $A$ to domain $B$  while retaining the semantics (the meaning of semantic here is task-specific) we present an image generator network which is composed of five main parts: 1) Image generator from domain $A$ to B: \(G_A\); 2) Image generator from domain $B$ to A: \(G_B\); 3) Adversarial discriminator \(D_A\) to distinguish between generated images \(G_B(x^B)\) and real domain $A$ images \(x^A\); 4) Adversarial discriminator \(D_B\) to distinguish between generated images \(G_A(x^A)\) and real domain $B$  images \(x^B\); 5) A task network \(T^B\) in domain $B$ , in which the inputs are generated from domain $B$ images \(G_A(x^A)\) and the target labels are corresponding ground-truth labels \(y_i^{A}\in{Y}^{A}\). Parts 1-4 are the same as CycleGAN work from (Zhu \etal \cite{Zhu2017}), and are used to generate realistic fake images. Part 5 is the key to retaining semantic consistency between real image \(x^A\) and generated image \(G_A(x^A)\). The overview of the proposed method is shown in Figure \ref{fig:network}.

In terms of training this network, there are several losses that need to be optimized; among these losses 1-3 are the same as (Zhu \etal \cite{Zhu2017}).

1)	Adversarial loss: The adversarial loss (Goodfellow \etal \cite{Goodfellow2014}) is applied to both generators and their corresponding discriminators. The objective of having adversarial loss is to train the generator network to generate visually realistic images in the target domain. 
For the generator from domain $A$ to B, the loss is as follows
\begin{equation} \label{eq:1}
\begin{split}
L_{GAN}\left(G_{A},{D}_{B},{x}^{A},x^B\right)={E}_{{x}^{B}}\left[\mathrm{log}\,{D}_{B}\left(x^B\right)\right]+ \\
{E}_{{x}^{A}}\left[\mathrm{log}\left(\mathbf{1}-{D}_{B}\left(G_{A}\left(x^{A}\right)\right)\right)\right],
\end{split}
\end{equation}

\begin{equation} \label{eq:2}
\begin{split}
L_{GAN}\left(G_{B},{D}_{A},{x}^{B},x^{A}\right)={E}_{{x}^{A}}\left[\mathrm{log}\,{D}_{A}\left(x^{A}\right)\right]+ \\
{E}_{{x}^{B}}\left[\mathrm{log}\left(\mathbf{1}-{D}_{A}\left(G_{B}\left(x^{B}\right)\right)\right)\right].
\end{split}
\end{equation}

2)	Cycle consistency loss: There are an infinite number of possible image mappings from one domain to the other while matching the target domain distributions. To constrain the space of this mapping function, Zhu \etal \cite{Zhu2017} introduced cycle consistency loss which forces the image translation cycle to return the input image back to the original image \((x^{A}\ \approx\ \ G_{B}\left(G_{A}\left(x^{A}\right)\right)\) and \(x^{B}\ \approx\ \ G_{A}\left(G_{B}\left(x^{B}\right)\right)\)). The cycle consistency loss is expressed as
\begin{equation} \label{eq:3}
\begin{split}
L_{Cycle}\left(G_{A},G_{B}\right)={E}_{{x}^{A}}\|G_{B}(G_{A}(x^{A})) - x^{A}\|_1 + \\ {E}_{{x}^{B}}\|G_{A}(G_{B}(x^{B})) - x^{B}\|_1.
\end{split}
\end{equation}

3)	Identity loss: To constrain the image generator to preserve color information between the input and output, an identity loss is applied. The identity loss was first introduced by Taigman \etal \cite{Taigman2017}, and is defined as 
\begin{equation} \label{eq:4}
\begin{split}
L_{Identity}\left(G_{A},G_{B}\right)={E}_{{x}^{B}}\|G_{A}(x^{B}) - x^{B}\|_1 + \\
{E}_{{x}^{A}}\|G_{B}(x^{A}) - x^{A}\|_1.
\end{split}
\end{equation}

\noindent The intuition behind addition of the identity loss is to enforce the generator to be an identity mapping when images from the source domain are fed into the generator.

4)	Task-specific semantic constraint loss: Using losses 1-3 we can train a pair of generators that generate visually realistic images in the target domains. However, aside from the existence of the cycle consistency loss, the semantics are not specifically constrained after translation. The semantics, especially the detailed spatial semantics such as the position and size of the fruit, are prone to change, which makes the generated labels unusable in terms of domain adaptation when localization is required. To overcome this limitation, we use a task-specific semantic constraint loss: Given a task model in domain $B$ , \(T^B\), the \(T^B\) prediction result of generated image \(G_{A}\left(x^{A}\right)\) should be identical to \(x^A\)’s  ground truth label \(y^A\). During backpropagation and parameter updates, the weights in \(T^B\) are fixed, the gradient is passed into the generator \(G_{A}\) to encourage it to generate images that can let \(T^B\) generate more accurate predictions. We find this is a very data-efficient way to extract knowledge from \(T^B\) to help \(G_{A}\) generate semantically consistent translated images. This loss is referred as \(L_{task}\left(G_{A}\right)\) and its specific form depends on the task and task model (e.g., YOLOv3 has its specific loss). The only requirement of this loss and the task is that the task loss is differentiable. 

Combining all the losses above, the full objective function is given by:
\begin{equation} \label{eq:5}
\begin{split}
L\left(G_{A},D_{B},G_{B},D_{A}\right)=L_{GAN}\left(G_{A},D_{B},x^{A},{x}^{B}\right) +\\ L_{GAN}\left(G_{B},D_{A},x^{B},x^{A}\right) +  \lambda_cL_{Cycle}\left(G_{A},G_{B}\right) + \\ \lambda_i{\ L}_{Identity}\left(G_{A},G_{B}\right) +  \lambda_tL_{task}\left(G_{A}\right),\ 
\end{split}
\end{equation}
where \(\lambda_c,\ \lambda_i,\ \lambda_t\) are relative weights of the cycle consistency loss, identity loss, and task specific semantic constraint loss. When \(\lambda_t\ =\ 0\) this method collapses to the original CycleGAN method.

\subsubsection{Fine-tune the task network using generated images}
In the last step, we get a semantically consistent image generator \(G_{A}\). Applying \(G_{A}\) to all the domain $A$ data \({x}^{A}\) we can get the same number of generated data \(G_{A}({x}^{A})\) with labels that correspond with \(y^A\). Using this generated data, we can further train the task network to improve performance in domain $B$ . 

\subsection{Network architectures}
The generator and discriminators \(G_{A},D_{B},G_{B},D_{A}\) are the same as those in Zhu \etal \cite{Zhu2017}. The task network \(T^B\) is YOLOv3-tiny which is a very light-weighted model proposed by Redmon \etal \cite{Redmon2018}. The main reason for choosing YOLOv3-tiny as the task network \(T^B\) is to decrease the training time, but other differentiable task networks could be used as well.

\section{Experiments and Results}
\subsection{3D synthetic source domain}
One special domain is the 3D synthetic domain, where the images are generated using a rendering engine instead of collected in the real world. The benefit of generating images from a rendering engine is that the ground-truth labels are known and easily extracted. The synthetic domain can be treated as a domain with “infinite” labeled images. On the other hand, no matter how realistic the synthetic images are, there is always a domain gap between the synthetic image and the real-world domain where the model needs to be applied. Moreover, the level of photorealism increases rendering time which reduces the scalability of synthetic image generation within agricultural applications. This gap lead to model performance degradation in the target domain.  

In this work, we used the open source Helios 3D Plant and Environmental Biophysical Modeling Framework of Bailey \cite{Bailey2019} to generate synthetic grape images. Using Helios, we generated 500 synthetic vineyard images that spanned a range of geometric canopy parameters, trellis types, and camera positions. Bounding box labels for grape clusters were generated using a custom Helios plugin. Importantly, Helios can be used to parametrically generate 3D geometries for a wide range of crop types which can then be used to create synthetically labeled data, including those for object detection, semantic segmentation, or instance segmentation. 

\subsection{Real world target domain}
We have two target domains in this work, one we called day domain and the other one called night domain. 

\subsubsection{Day domain}
The day domain data were collected in the California Central Valley using a GoPro camera in Summer 2019 during the daytime. 3065 images are in this dataset and we labeled 100 images; among them 25 images are always used for the test set to evaluate the model performance in this domain. Example images in the day domain are shown in Figure \ref{fig:domain}.

\subsubsection{Night domain}
\begin{figure}[h]
\begin{center}
   \includegraphics[width=1.0\linewidth]{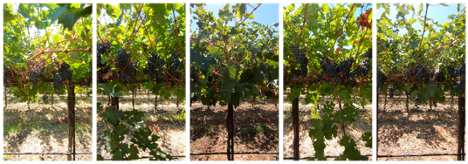}
   \includegraphics[width=1.0\linewidth]{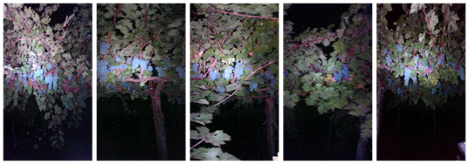}
\end{center}
   \caption{Top: example day domain real images; bottom: example night domain real images.}
\label{fig:domain}
\end{figure}

The night domain data were collected at the same location as the day domain using an Intel RealSense D435i camera with a custom lighting system in Summer 2020 during nighttime. 800 images are in this dataset and we labeled 150 images; among them 24 are always used for the test set to evaluate the model performance in this domain. Example images in the night domain are shown in Figure \ref{fig:domain}.

\subsection{Experimental Design}

The main idea of our work is to utilize a 3D crop model and a GAN model to reduce the need for labeling in a new domain. The task we choose here is grape detection and the detection model we use is YOLOv3 with the tiny-backbone (Redmon and Farhadi \cite{Redmon2018}).  We have test datasets for each target domain that do not engage in the model training but are just used for evaluating the final model performance.

To evaluate how using the combined 3D crop model and GAN approach affects data efficiency, we pre-trained an object detection model $T$ with the synthetic images (3D crop model generated) and the generated labels. We evaluate the performance of the pre-trained model $T$, the performance of the model $T$ after further fine-tuning using $N$ labeled target domain real images, and the performance of the model $T$ after fine-tuning using $N$ labeled target domain real images and GAN generated images with source domain labels.

We choose $N$ = 2, 5, 10, 15, 20, 30, 40, 50. Among them, 80\% of the labeled images are used for training ($a$) and the remaining 20\% of images are used for validation ($b$) (at least 1 image in each set). The best performing model on the validation set during the training is selected. The GAN for each experiment also uses the same fine-tuned model $T$ using $N$ labeled target domain images; no additional labeled images are introduced into training the GAN. We also evaluated the performance of the model if we use only CycleGAN to generate images and fine-tune the model $T$ using these generated images. We use AP (Average Precision, see Everingham \etal \cite{Everingham2010} for detailed definition of AP) at 0.3 and 0.5 IoU (Intersection over Union) as model performance metrics. 

\subsubsection{Generate images using semantically constrained GAN}
\begin{figure*}[h]
\begin{center}
   \includegraphics[width=0.75\linewidth]{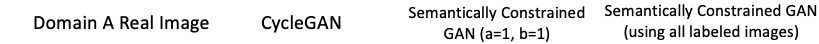}
   \includegraphics[width=0.75\linewidth]{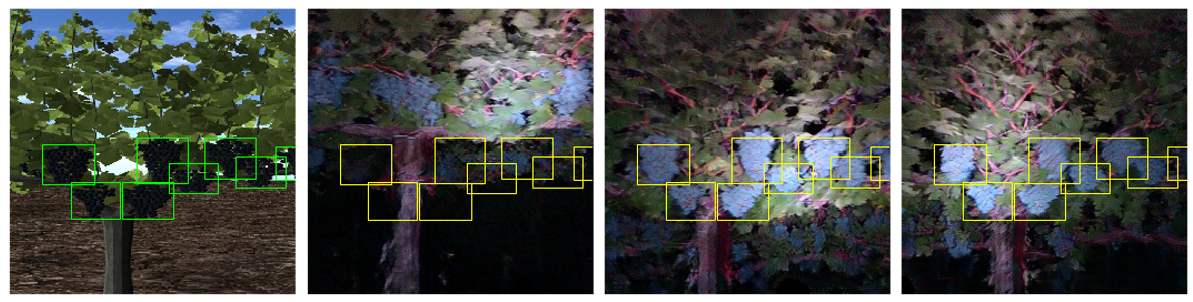}
   \includegraphics[width=0.75\linewidth]{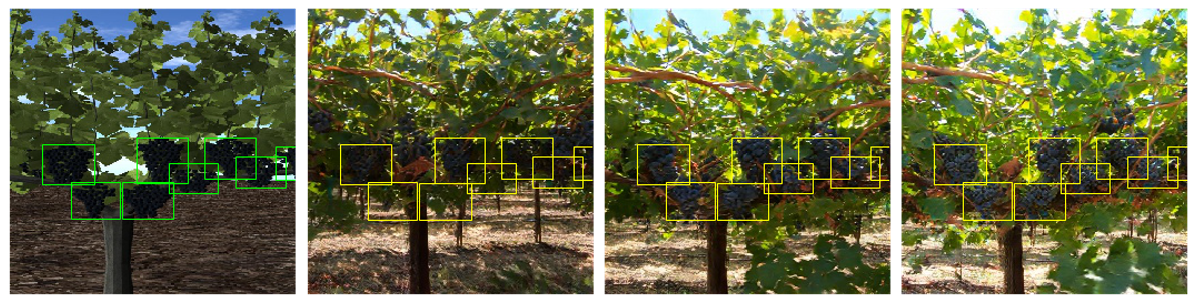}
   \includegraphics[width=0.75\linewidth]{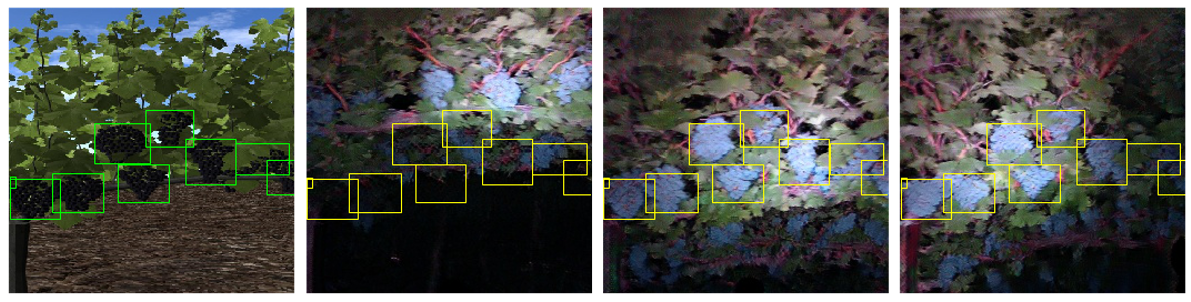}
   \includegraphics[width=0.75\linewidth]{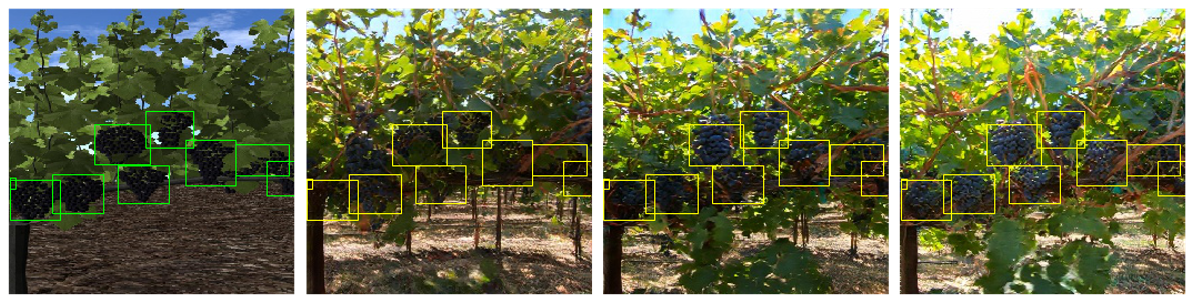}
\end{center}
   \caption{Example GAN generated images (randomly selected). The first column is randomly selected source domain images with the ground truth generated labels. The 2 – 4 columns are generated images with the projected labels in the yellow box (same as the label in the first column, just for visualization purposes). $a$ is the number of labeled target domain image for train, $b$ is the number of labeled target domain image for validation.}
\label{fig:generated-image}
\end{figure*}

To help better understand the quality of generated images using the semantically constrained GANs, a set of random results is shown in Figure \ref{fig:generated-image}. The source images in domain $A$ are randomly selected, each two rows are using the same source image from the first column and translate into different target domains. From the generated images, we can see that the baseline CycleGAN and the semantically constrained CycleGAN models can generate visually realistic images. However, the baseline CycleGAN has trouble in generating images with the grapes in the same location as the source synthetic images. This "positional drift" problem is more significant in the generated night domain images than the day domain images generation. The main reason for this drift is that the CycleGAN network is not provided information to learn what a grape is, and the domain gap between the night domain to the synthetic domain is relatively large. Using the semantically constrained GAN, even when the task constrained network is trained with only 1 labeled image and validated on only 1 labeled image, the generated image can be very well semantically constrained, in terms of grape position and size. Also, a single source domain 3D rendered image can be generated for two different real-world domain images using two generators, and both the generated images show the same grape distribution as the 3D rendered image.

\subsubsection{Fruit detection performance}
We first trained a task network only using synthetic 3D grape model images (using 345 train and 74 validation). The results of applying this model into two target real-world domains are shown in Table \ref{table:2} and the baseline methods' results are shown in Table \ref{table:1}. The performance of the direct synthetic to real model transfer is shown in the first rows labeled “Synthetic pre-trained”. We also applied the CycleGAN method using the synthetic–night and synthetic–day images, generated images in target domains, and fine-tuned the pre-trained task model on these generated images (validate on 20\% of the generated images). The results are shown in the second rows labeled “CycleGAN”.  The columns of “Only fine-tuned” contain the results of fine-tuning the pre-trained task mode using \(a\) labeled target domain train images, and selecting model based on \(b\) labeled target domain valid images. The performance of the task networks refined using our Semantically Constrained GAN is shown in the “GAN refined” columns. As we can see from the results, the direct application of a model pre-trained on the 3D synthetic domain to the real domain can result in relatively poor performance since the real domains are different than the 3D synthetic domain. Especially for the night domain, the pre-trained model has almost no ability to predict grape locations. One na\"{i}ve domain adaptation method is using CycleGAN to generate target domain images, assuming the labels are the same as the source images, and further training the pre-trained model on these generated images and labels. The experiment shows that this approach will not lead to performance improvement, and can even lead to a decrease in performance in the day domain. The main reason is that the generated images do not always keep the grape clusters at the same location and thus the source labels no longer valid. Another domain adaptation method is to use some labeled images in the target domain to fine tune the pre-trained network. This classical method is still very promising and leads to a significant increase in model performance even using 1 labeled training image. The performance of the model increases with increasing number of labeled target domain images involved. Our method can further improve the data efficiency upon the fine-tuning using the same labeled target domain images, especially at a very low number of labeled target domain images. 
\begin{table*}[]
\begin{center}
\begin{tabular}{|c|c|c|c|c|c|c|c|}
\hline
\multirow{2}{*}{Baseline   methods} & \multirow{2}{*}{\begin{tabular}[c]{@{}c@{}}Train  \\ num a\end{tabular}} & \multirow{2}{*}{\begin{tabular}[c]{@{}c@{}}Valid  \\ num b\end{tabular}} & \multirow{2}{*}{\begin{tabular}[c]{@{}c@{}}Total \\ num k\end{tabular}} & \multicolumn{2}{c|}{\textbf{Synthetic to Day Domain}} & \multicolumn{2}{c|}{\textbf{Synthetic to Night Domain}} \\ \cline{5-8} 
 &  &  &  & \begin{tabular}[c]{@{}c@{}}AP@\\ IOU0.3\end{tabular} & \begin{tabular}[c]{@{}c@{}}AP@\\ IOU0.5\end{tabular} & \begin{tabular}[c]{@{}c@{}}AP@\\ IOU0.3\end{tabular} & \begin{tabular}[c]{@{}c@{}}AP@\\ IOU0.5\end{tabular} \\ \hline
Synthetic Pre-trained & 0 & 0 & 0 & 27.8 & 13.2 & 0.0 & 0.8 \\ \hline
Cycle GAN & 0 & 0 & 0 & 10.3 & 2.1 & 0.1 & 0.0 \\ \hline
\end{tabular}
\end{center}
\caption{Grape detection results using baseline domain adaptation methods. Average precision (AP) numbers are percentage. Synthetic Pre-trained means only using synthetically generated image to train the model. Cycle GAN means using Cycle GAN generated images to fine-tune the pre-trained model. All models are evaluated on the same test dataset as Table \ref{table:2}.}
\label{table:1}
\end{table*}

\begin{table*}[]
\begin{center}
\begin{tabular}{|c|c|c|c|c|c|c|c|c|c|c|}
\hline
\multirow{3}{*}{\begin{tabular}[c]{@{}c@{}}Train \\  num a\end{tabular}} & \multirow{3}{*}{\begin{tabular}[c]{@{}c@{}}Valid  \\ num b\end{tabular}} & \multirow{3}{*}{\begin{tabular}[c]{@{}c@{}}Total \\ num k\end{tabular}} & \multicolumn{4}{c|}{\textbf{Synthetic to Day Domain}} & \multicolumn{4}{c|}{\textbf{Synthetic to Night Domain}} \\ \cline{4-11} 
 &  &  & \multicolumn{2}{c|}{fine-tuned} & \multicolumn{2}{c|}{\begin{tabular}[c]{@{}c@{}}SemGAN   +\\ fine-tuned\end{tabular}} & \multicolumn{2}{c|}{fine-tuned} & \multicolumn{2}{c|}{\begin{tabular}[c]{@{}c@{}}SemGAN   +\\        fine-tunned\end{tabular}} \\ \cline{4-11} 
 &  &  & \multicolumn{1}{l|}{\begin{tabular}[c]{@{}l@{}}AP@\\ IOU0.3\end{tabular}} & \multicolumn{1}{l|}{\begin{tabular}[c]{@{}l@{}}AP@\\ IOU0.5\end{tabular}} & \multicolumn{1}{l|}{\begin{tabular}[c]{@{}l@{}}AP@\\ IOU0.3\end{tabular}} & \multicolumn{1}{l|}{\begin{tabular}[c]{@{}l@{}}AP@\\ IOU0.5\end{tabular}} & \multicolumn{1}{l|}{\begin{tabular}[c]{@{}l@{}}AP@\\ IOU0.3\end{tabular}} & \multicolumn{1}{l|}{\begin{tabular}[c]{@{}l@{}}AP@\\ IOU0.5\end{tabular}} & \multicolumn{1}{l|}{\begin{tabular}[c]{@{}l@{}}AP@\\ IOU0.3\end{tabular}} & \multicolumn{1}{l|}{\begin{tabular}[c]{@{}l@{}}AP@\\ IOU0.5\end{tabular}} \\ \hline
1 & 1 & 2 & 37.4 & 16.4 & \textbf{51.0} & \textbf{23.4} & 32.6 & 8.3 & \textbf{38.3} & \textbf{12.1} \\ \hline
4 & 1 & 5 & 49.7 & 21.8 & \textbf{51.8} & \textbf{23.6} & 38.2 & 10.8 & 37.8 & \textbf{12.2} \\ \hline
8 & 1 & 9 & 39.8 & 16.5 & \textbf{55.7} & \textbf{26.5} & 35.7 & 9.3 & \textbf{38.2} & \textbf{13.0} \\ \hline
12 & 2 & 14 & 52.1 & 26.6 & \textbf{54.7} & 26.6 & 43.0 & 13.9 & \textbf{45.1} & 12.8 \\ \hline
16 & 3 & 19 & 51.6 & 23.7 & \textbf{57.9} & \textbf{30.8} & 43.0 & 13.4 & \textbf{46.1} & \textbf{17.3} \\ \hline
24 & 6 & 30 & 56.3 & 26.5 & \textbf{57.6} & \textbf{28.5} & 45.2 & 17.5 & \textbf{48.2} & \textbf{20.2} \\ \hline
32 & 8 & 40 & 57.9 & 31.1 & \textbf{59.5} & \textbf{36.1} & 46.1 & 17.5 & \textbf{51.4} & \textbf{20.2} \\ \hline
40 & 10 & 50 & 57.4 & 31.7 & \textbf{63.9} & \textbf{36.4} & 49.7 & 19.3 & \textbf{50.5} & \textbf{20.6} \\ \hline
98(all) & 28(all) & 126 & / & / & \textbf{/} & \textbf{/} & 52.8 & 22.8 & \textbf{56.0} & \textbf{26.7} \\ \hline
58(all) & 15(all) & 73 & 61.3 & 37.0 & \textbf{61.7} & \textbf{37.2} & / & / & \textbf{/} & \textbf{/} \\ \hline
\end{tabular}
\end{center}
\caption{Grape detection results.Average precision (AP) numbers are percentage. "/" means not applicable. "SemGAN + fine-tuned" columns are the results using the semantically constrained cycle GAN generated images to fine-tune the detection network. The "fine-tuned" results are just using the \(a\) labeled train images in the target domain to fine-tune the detection network. Network are selected using corresponding \(b\) labeled valid images. For each domain, all models are evaluated on the same test dataset.}
\label{table:2}
\end{table*}

\section{Discussion and Future Work}
To apply deep learning-based AI models in agricultural and plant environments, we need to overcome the problem of insufficient labeled data and massive variability (e.g., plant appearance, horticultural practice, seasonal differences, lighting differences). It is labor and cost-intensive to manually label images in the broad range of scenarios that can be encountered in agricultural production environments, and doing so will hinder the large scale adoption of deep learning model deployment in agricultural production. To solve these problems and make deep learning model deployment more feasible in new agricultural environments, we proposed a semantically constrained GAN. We presented a training pipeline for this network and used the generated images to improve task model performance in a new domain, i.e., fruit detection. The results in this paper showed that by using a semantically constrained GAN we can generate very realistic day and night grapevine images from 3D rendering images while retaining grape position and size. The generated images can be used to further train the task network and improve the task network performance in the target domain which can surpass the vanilla fine-tuning results, especially with a low number of labeled images. 

Many interesting questions remain to be answered following this research. 1) Using this method, we successfully constrained grape position and geometry, but other parts of the images are unconstrained (e.g. foliage, trunks, sky, etc.). The reason that they are not constrained is that the task-constrained network is only designed to identify grapes. It would be interesting to see if the task-constrained network can identify and constrain multi-class objects, or even constrain the whole scene semantics by replacing the object detection task network with a semantic segmentation task network. 2) When further training using the GAN generated images, we did not include the generated images in the validation set, only the true labeled images (except when using CycleGAN, since there were no labeled images involved). However, adding GAN-generated images into the validation set to select the best model can also help to improve the overall model accuracy, especially when the labeled validation image number is low or even when no labeled validation images are included. Determining the best mixing ratio between GAN-generated images and labeled images in the validation set could further improve data efficiency. 3) The main GAN network architecture of this work is the same as the CycleGAN work except the task constrained network. While only adding the task constrained network already achieves good semantic consistency, there is some work that focuses on other ways to achieve semantic consistency such as Hoffman \etal \cite{Hoffman2018} and Chen \etal \cite{Chen2019}. It would be interesting to see how model performance changes when we utilize these semantic consistency methods.

\section{Acknowledgement}
This project was partly supported by the USDA AI Institute for Next Generation Food Systems (AIFS), USDA award number 2020-67021-32855, and by the NSF funded Center for Data Science and Artificial Intelligence award number 1934568. Brian N. Bailey was supported by USDA National Institute of Food and Agriculture Hatch project 1013396. We would also like to thank Hamid Kamangir and Kaustubh Deshpande for support on synthetic data generation. 
{\small
\bibliographystyle{ieee_fullname}
\bibliography{SemGan}
}

\end{document}